\documentclass[letterpaper, 10 pt, conference]{ieeeconf}  % Comment this line out if you need a4paper

\IEEEoverridecommandlockouts                              % This command is only needed if 
\usepackage{graphicx}
\usepackage{algorithm}
\usepackage{algpseudocode}
\usepackage{subcaption}
\usepackage{hyperref}
\usepackage[english]{babel}
\usepackage{tabularx}
\usepackage{booktabs}
\usepackage{multirow}
\usepackage{amsmath}
\usepackage[inkscapelatex=false]{svg}
\usepackage{cleveref}
\usepackage{stfloats}
\usepackage{float}
\title{\LARGE \bf
Scene Graph-Driven Haptic Feedback for Safety Enhancement in Robotic Ophthalmic Surgery via Physically Simulated iOCT
}
\author{Danial Arbabi$^{1}$, Korab Hoxha$^{1}$, Angelo Henriques$^{1}$, Mirza Imamovic$^{1,2,3}$, and M. Ali Nasseri$^{1,2,4}$% <-this % stops a space
\thanks{*This paper is supported by the state of Bavaria through Bayerische Forschungsstiftung
(BFS) under Grant AZ-1592-23-ForNeRo.
}% <-this % stops a space
\thanks{$^{1}$School of Medicine and Health, Department of Ophthalmology, 
TUM University Hospital, Technical University of Munich (TUM), Munich, Germany}%
\thanks{$^{2}$Munich Institute of Robotics and Machine Intelligence (MIRMI), 
Technical University of Munich (TUM), Munich, Germany}%
\thanks{$^{3}$Chair of Ergonomics, TUM School of Engineering and Design, 
Technical University of Munich (TUM), Munich, Germany}%
\thanks{$^{4}$Department of Biomedical Engineering, 
University of Alberta, Edmonton, AB, Canada}%
\thanks{\copyright~2026 IEEE. Personal use of this material is permitted. Permission from IEEE 
must be obtained for all other uses, in any current or future media, including 
reprinting/republishing this material for advertising or promotional purposes, creating new 
collective works, for resale or redistribution to servers or lists, or reuse of any copyrighted 
component of this work in other works.}%
}

\usepackage{xcolor}
\begin{document}

\maketitle
\thispagestyle{empty}
\pagestyle{empty}

\begin{abstract}

Robotic ophthalmic surgery offers high precision but introduces a "sensory gap" by decoupling the surgeon from their instrument, resulting in a loss of tactile feedback.
This paper presents a novel haptic feedback system for subretinal injection tasks leveraging Scene Graphs (SG).
The system bridges the sensory gap by analyzing a physically simulated intraoperative Optical Coherence Tomography (iOCT) feed to construct a real-time surgical SG.
The SG serves as a semantic abstraction layer for the surgical scene, which is then utilized by a deterministic, rule-based engine to generate state-dependent haptic feedback on a robotic input device.
The system was evaluated in a user study (N=16) using an anthropomorphic head phantom and a custom-built surgical robot.
Results demonstrate that the SG-driven haptic feedback improved surgical precision, reducing needle alignment error by $14\%$ ($p=0.044$) and improving System Usability Scale (SUS) scores by $8\%$ ($p=0.015$), while maintaining comparable task completion times.
A needle trajectory analysis revealed the emergence of a safer "Align-then-Approach" strategy, in which our haptic negative reinforcement prompted users to fine-tune the tool's trajectory before approaching the retinal target.
This work suggests that SGs can effectively serve as the direct computational foundation for real-time, safety-enhancing context-aware haptic feedback in robotic microsurgery.
\end{abstract}

\section{INTRODUCTION}
    % \begin{figure}[t]
%     \centering    \includegraphics[width=1.0\columnwidth]{pictures/Subretinal_Injection_Schematic.pdf} 
%     \caption{
%     % Schematic of a subretinal injection. 
%     % The goal is to deliver a therapeutic agent into the space between the neurosensory retina and the retinal pigment epithelium (RPE)~\cite{labbate_biomechanical_2024}.
%     % The medication forms a localized retinal detachment (bleb).
%     % This procedure requires high-precision depth control to pierce the neurosensory retina without damaging the underlying layers.
%     Schematic of subretinal injection. The needle targets the space between the neurosensory retina and the RPE~\cite{labbate_biomechanical_2024}.
%     Concept adapted from~\cite{irigoyen_subretinal_2022}. %\textcolor{red}{Better not fundamentally change the layout of the images and tables. It kind of makes sense as it is now :)}
%     }
%     \vspace{-12pt}
%     \label{fig:clinical_schematic}
% \end{figure}

Subretinal injection is a surgical procedure used to treat pathologies such as age-related macular degeneration (AMD)~\cite{tripepi_role_2023}.
The procedure requires sub-millimeter precision~\cite{singh_overcoming_2022}, often exceeding the physiological limits of human surgeons due to hand tremor.
Thus, robotic systems have been introduced~\cite{dimaio_da_2011, nasseri_introduction_2013} to filter and scale motion, enhancing the precision and improving safety.

However, current robotic systems introduce a sensory gap~\cite{el_rassi_review_2020},
effectively blocking direct tactile feedback from the instrument.
% By mechanically decoupling the surgeon from their instruments and thus also eliminating tremor transfer, the system
%the surgeon loses the ability to perceive interaction forces, used to judge instrument-tissue interactions, directly.
Furthermore, interaction forces in subretinal injections drop below the threshold of human perception, shifting reliance entirely to visual estimation and possibly increasing cognitive load~\cite{enayati_robotic_2018}.
% This leaves surgeons to rely entirely on other information sources and senses, which can increase cognitive load~\cite{?}.

To address this, strategies using haptic feedback~\cite{bergholz_benefits_2023} have been proposed.
While force sensors, used to measure forces acting on the instruments and replicate them to an input device, often require complex hardware changes, using Visual-to-Haptic feedback offers an alternative. This approach allows computer vision to infer geometry and forces via existing infrastructure, such as iOCT.

% To address this challenge, various substitution methods have been proposed, like robotic systems mirroring forces acting on the instruments to the surgeon's robotic input device.
% Sadly, this force replication strategy is rendered inefficient, as the tactile forces involved fall below the threshold of human perception~\cite{?}.

In this work, we propose a novel framework that leverages SGs as the direct foundation for a real-time, context-aware feedback system, connecting visual perception and haptic feedback.
Unlike raw sensor data and distance metrics, SGs provide a structured representation of a surgical scene~\cite{HENRIQUES2026104083}, which we then combine with a rule-based engine to infer context-aware vibrotactile cues for the surgeon. This system aims to synthesize kinesthetic forces from semantic vibrotactile cues.

This paper makes the following contributions:
\begin{enumerate}
    \item A perception pipeline that builds a real-time SG from a simulated iOCT by having the side view from a physical eye phantom.
    \item A rule-based inference engine mapping SG relations to specific haptic feedback patterns.
    %Quantitative results evaluating the system and demonstrating improved safety and usability through a user study with the MAPS' custom surgical robot.
\end{enumerate}
The results of this work were validated in a user study (N=16) on an existing surgical robotic platform.

\section{Background and Related Work}
    \subsection{Clinical Motivation: Subretinal Injection}
        \begin{figure}[t]
    \centering    \includegraphics[width=1.0\columnwidth]{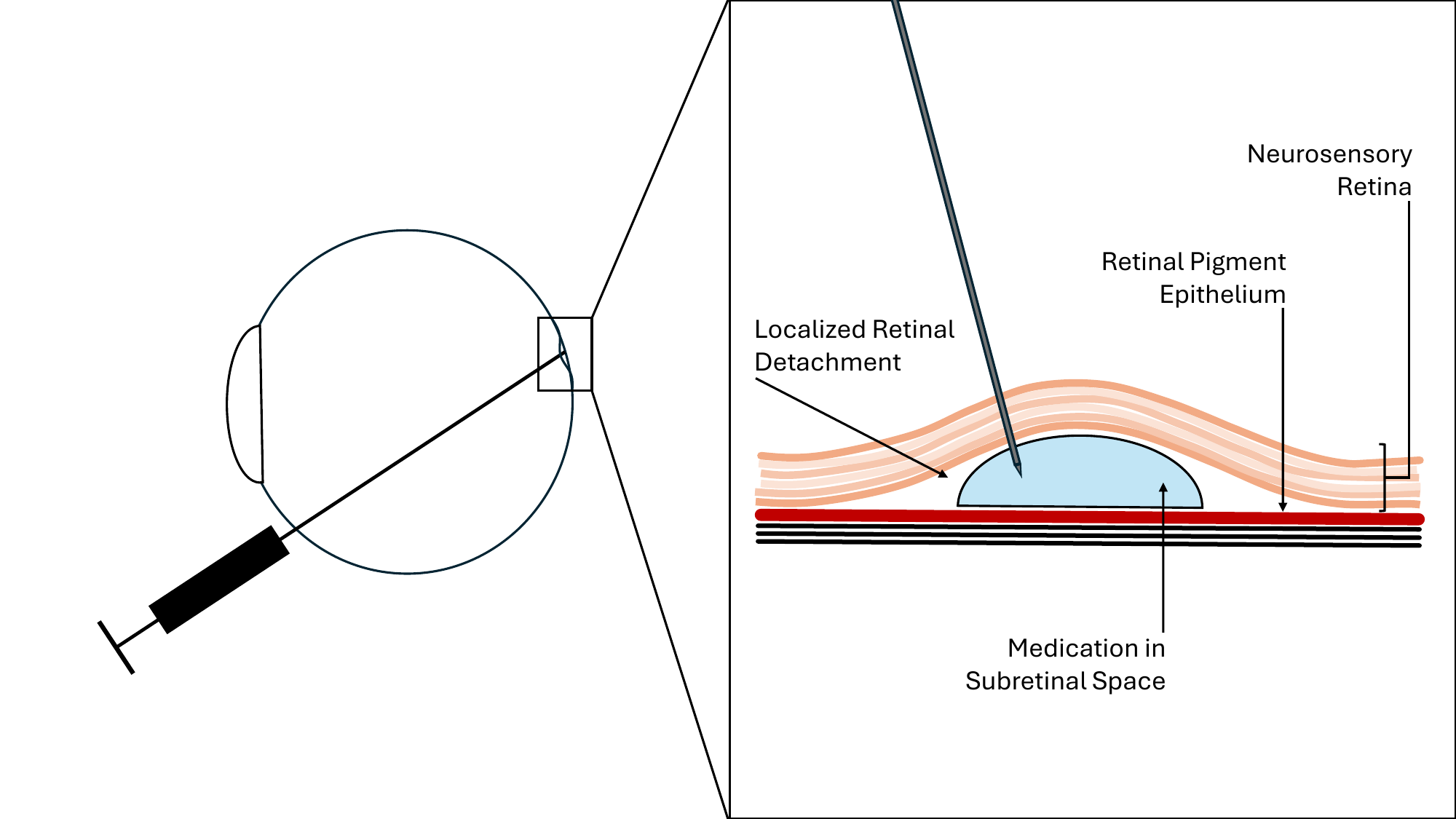} 
    \caption{
    % Schematic of a subretinal injection. 
    % The goal is to deliver a therapeutic agent into the space between the neurosensory retina and the retinal pigment epithelium (RPE)~\cite{labbate_biomechanical_2024}.
    % The medication forms a localized retinal detachment (bleb).
    % This procedure requires high-precision depth control to pierce the neurosensory retina without damaging the underlying layers.
    Schematic of subretinal injection. The needle targets the subretinal space between the neurosensory retina and the RPE~\cite{labbate_biomechanical_2024}.
    Concept adapted from~\cite{irigoyen_subretinal_2022}. %\textcolor{red}{Better not fundamentally change the layout of the images and tables. It kind of makes sense as it is now :)}
    }
    \vspace{-12pt}
    \label{fig:clinical_schematic}
\end{figure}
A subretinal injection (\Cref{fig:clinical_schematic}) is an ophthalmic microsurgical procedure in which a controlled dose of medication is injected into predefined retinal layers, reaching photoreceptors and the retinal pigment epithelium (RPE)~\cite{xue_technique_2017}.
Used for treating pathologies such as submacular hemorrhages, macular folds, and AMD, as well as administering gene- and stem-cell cargos~\cite{tripepi_role_2023}, the procedure demands extreme precision.
%which is a "major cause of blindness worldwide"~\cite{lim_age-related_2012}, and are the preferred method for treating inherited retinal diseases through gene therapy.
% The procedure is physically and technically demanding.
Surgeons are required to navigate a needle through the vitreous cavity and penetrate the neurosensory retina without damaging the underlying layers, often assisted by iOCT~\cite{tripepi_role_2023}.
% To assist with this delicate maneuver, iOCT can be focused on the injection position to precisely monitor the separation between the retina and the RPE below, thereby keeping overstretching within limits~\cite{tripepi_role_2023}.
However, physiological hand tremor and drift often exceed the safety bounds of micrometer-scale retinal tissue manipulation~\cite{singh_overcoming_2022}.
% despite the use of better visualization techniques and equipment, the aspect of manipulating this micrometer-thick layer remains challenging.
% The physiological hand tremor and drift often exceed the safety bounds of retinal tissue manipulation~\cite{singh_overcoming_2022}.

% \begin{figure}[h]
%     \centering
%     \begin{subfigure}[t]{0.48\columnwidth}
%         \centering
%         \includegraphics[width=\linewidth]{pictures/960px-Eye_disease_simulation,_normal_vision.jpg}
%         \caption{Normal vision~\cite{health_normal_nodate}.}
%         \label{fig:normal_eye}
%     \end{subfigure}
%     \hfill
%     \begin{subfigure}[t]{0.48\columnwidth}
%         \centering
%         \includegraphics[width=\linewidth]{pictures/amd_combined_2.png}
%         \caption{Simulated vision with AMD~\cite{health_scene_nodate}.}
%         \label{fig:AMD_eye}
%     \end{subfigure}
%     \caption{}
%     \label{}
% \end{figure}

% \begin{figure}[t]
%     \centering
%     \includegraphics[width=0.85\columnwidth]{pictures/amd_combined_2.png}
%     \caption{Comparison between normal vision (left) and the same scene as it may be perceived by a person with AMD (right).}
%     \label{fig:AMD_comparison}
% \end{figure}
    \subsection{Robotic Assistance and the Sensory Gap}
        
Teleoperated robotic platforms, such as the da Vinci Surgical System~\cite{dimaio_da_2011}, PRECEYES~\cite{ramamurthy_robotics_2022}, and the system used in this work~\cite{nasseri_introduction_2013}, enhance precision by filtering tremors and scaling motion. 
% down to sub-millimeter levels, thereby increasing precision.
However, most of these systems separate the surgeon from the instruments, eliminating direct tactile feedback.
% feeling any tissue feedback through indirect contact with the surgical instruments.
% While performing manual surgery, faint haptic cues are used to evaluate tissue resistance and other forces.
This sensory gap can lead to unintentional tissue damage and a reduced ability to perform complex manipulations~\cite{el_rassi_review_2020}, thereby forcing reliance on visual cues and increasing cognitive load.
While force and haptic feedback can be used, often relying on force-sensing hardware~\cite{el_rassi_review_2020}, a visual-to-haptic approach offers a purely software-driven alternative to haptic feedback, leveraging already existing imaging infrastructure.   
% In plain robotic teleoperation, these cues are lost to the system, which can even lead to unintentional tissue damage and a lack of ability to perform highly complex manipulations~\cite{el_rassi_review_2020}. 
% Consequently, surgeons often rely entirely on other senses, utilizing visual feedback from a microscope or screen to determine tissue contact.
% This increases the difficulty of judging exact distances and may increase cognitive load.
% To address this issue, more systems are incorporating haptic force and feedback replication to enhance the surgeon's awareness of the surgical tool's surroundings.

    \subsection{Scene Graphs in Surgical Robotics}
        Unlike traditional Virtual Fixtures that rely on geometric constraints and force feedback, our approach utilizes a SG as a semantic layer to provide context-aware vibrotactile guidance. SGs have recently emerged as a powerful tool for decoding the complex semantics of the operating room, advancing beyond simple object detection to capture relational context \cite{HENRIQUES2026104083, Islam.2020, Nwoye.2022}. 
% While early works focused on static spatial configurations, the field has rapidly gravitated towards dynamic, temporal modeling of the surgical site.
While the current dominant application focuses on workflow recognition and automated safety assessment, using \textit{Action Triplet Recognition} (e.g., $\langle \text{instrument}, \text{action}, \text{tissue} \rangle$), most frameworks only function as passive observers, generating descriptive analytics for post-operative review or monitoring. 
% Currently, the dominant application is \textit{Action Triplet Recognition} (e.g., $\langle \text{instrument}, \text{action}, \text{tissue} \rangle$), which serves as a foundation for downstream tasks such as workflow recognition and automated safety assessment \cite{Nwoye.2022, Murali.2024}. 
% However, existing SSG frameworks typically function as passive observers, generating descriptive analytics for post-operative review or monitoring. 
Transforming this graph into active robotic control and guidance remains an open challenge~\cite{HENRIQUES2026104083}. 
We address this by utilizing the SG not for logging, but as a real-time state machine driving our closed-loop haptic feedback.
% The transition from descriptive scene understanding to ``actionable intelligence'', where the graph structure directly informs robotic control or active guidance, remains an open challenge. 
% In this work, we bridge this gap by utilizing the SSG not merely as a logging tool, but as the real-time state machine driving a closed-loop haptic feedback system.
% \vspace{-5pt}
        
\section{SYSTEM DESIGN}
    To prove the concept of this work, the proposed framework combines a compact imaging setup, a computer vision pipeline, and a semantic rule-based engine to provide context-aware haptic feedback 
% signals over ROS 2~\cite{ros2} 
to a robotic input device.
% The proposed framework combines a physically simulated workflow of an iOCT-guided robotic surgery.
% With a computer vision pipeline that feeds an SG into a semantic rule-based reasoning engine, relations within the graph, such as visual spatial data \textit{$\langle$needle\_tip, touching, retina$\rangle$}, are translated into intuitive haptic feedback. 
The system physically simulates iOCT-guided robotic surgery, with an anthropomorphic workspace.
While this optical setup serves as a low-cost proxy for iOCT geometry (\cref{fig:sg_vis}, left), the Scene Graph framework is designed to be agnostic to the input source, allowing future migration to clinical OCT feeds.
    
    \subsection{Anthropomorphic Phantom and Imaging Setup}
        
\begin{figure}
    \centering
    \includegraphics[width=0.45\columnwidth]{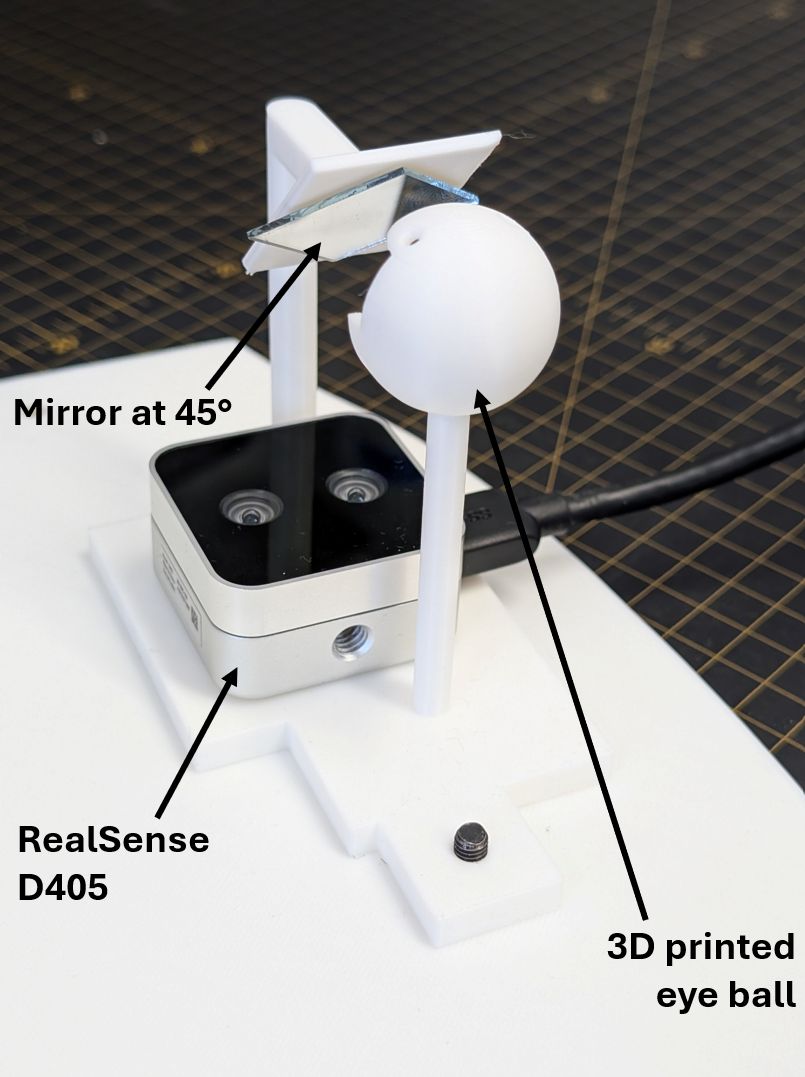}\hfill
    \includegraphics[width=0.45\columnwidth]{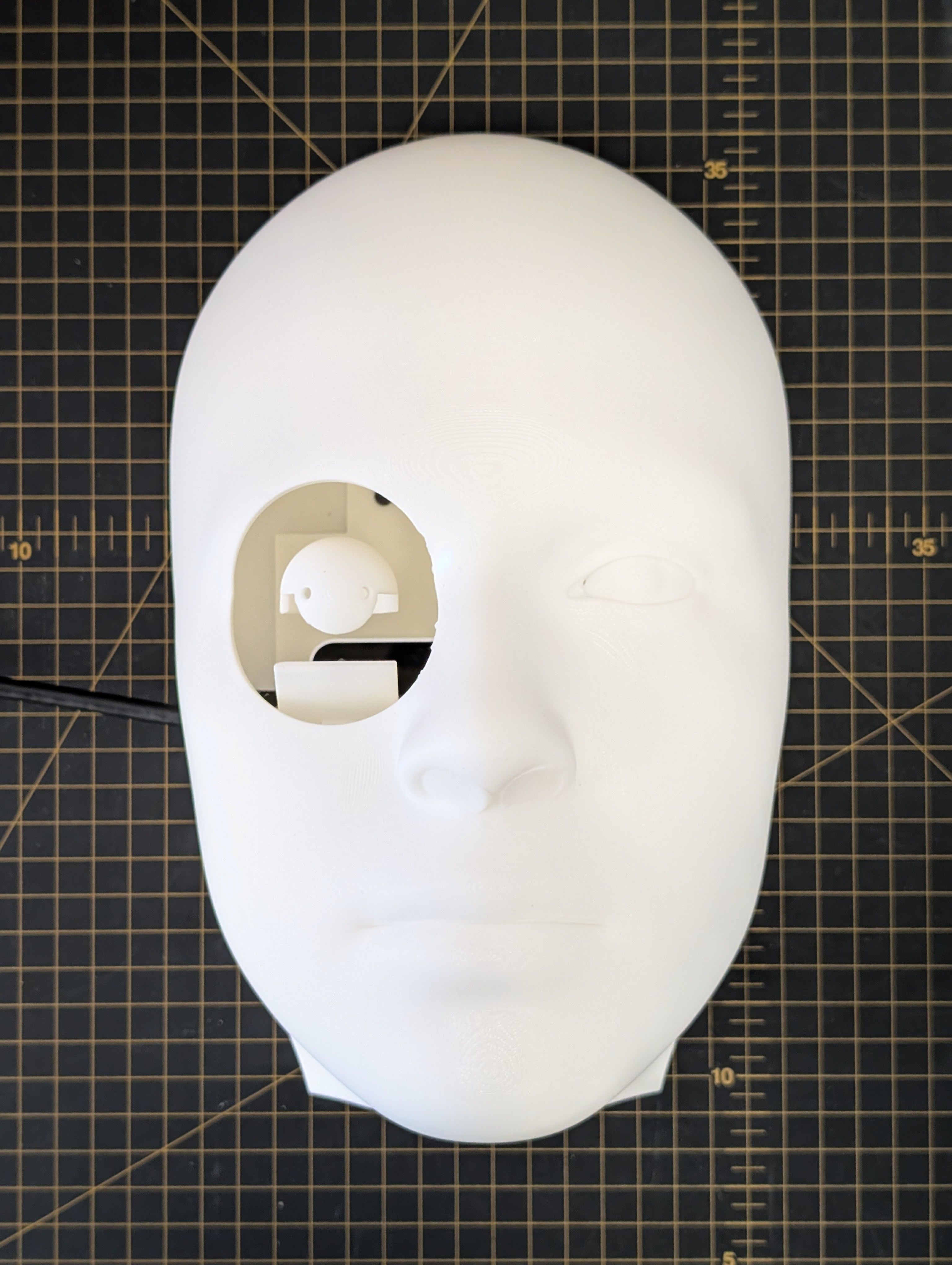}
    \caption{Imaging setup.
    \textbf{Left:} The internal optical module showing the D405 camera and 45° mirror.
    \textbf{Right:} The external view of the anthropomorphic face mask.
    }
    \vspace{-12pt}
    \label{fig:1}
\end{figure}

To evaluate our system under clinically relevant spatial constraints, we adapted a 3D-printed anthropomorphic phantom (\Cref{fig:1}).
Unlike standalone eye models, we integrate the general features of the human head to replicate generic facial constraints, such as the orbital rim and nasal bridge, that are encountered during ophthalmic procedures, which can sometimes limit the maneuverability of robotic arms.
The mirror-based optical module (\Cref{fig:1}, left) consists of a D405 RealSense camera, mounted horizontally, to capture a 2D cross-sectional view of the eye via a 45° mirror.
With this setup, the imaging hardware remains concealed within the skull geometry, capturing a cross-sectional view of the eye (1280 x 720 px @ 30 fps) and the instruments inside, while maintaining a realistic exterior form factor. 
% Integrating the imaging hardware without obstructing the robot's workspace and maintaining realism were key challenges in this setup.
% Addressing this, we developed a mirror-based optical module (\cref{fig:1}, left) embedded within the phantom to capture instrument movement inside the left eye.
The eye has a 24 mm diameter with a lateral cross-sectional cut and two holes for trocar placement and instrument insertion on the upper side of the sclera.
Since this work strictly focuses on iOCT B-Scan, the captured 2D images are sufficient for our iOCT approximation.
As the perception pipeline utilizes standard computer vision on a camera feed to simulate iOCT geometry, the SG acts as a dynamic modality-agnostic interface. Whether the nodes (e.g., 'Retina', 'Tool') are detected via intensity thresholding on a camera or U-Net segmentation on real iOCT, the downstream haptic inference engine remains unchanged.

\begin{figure}[h!]
    \centering
    % \vspace{-7pt}
    \includegraphics[width=0.922\columnwidth]{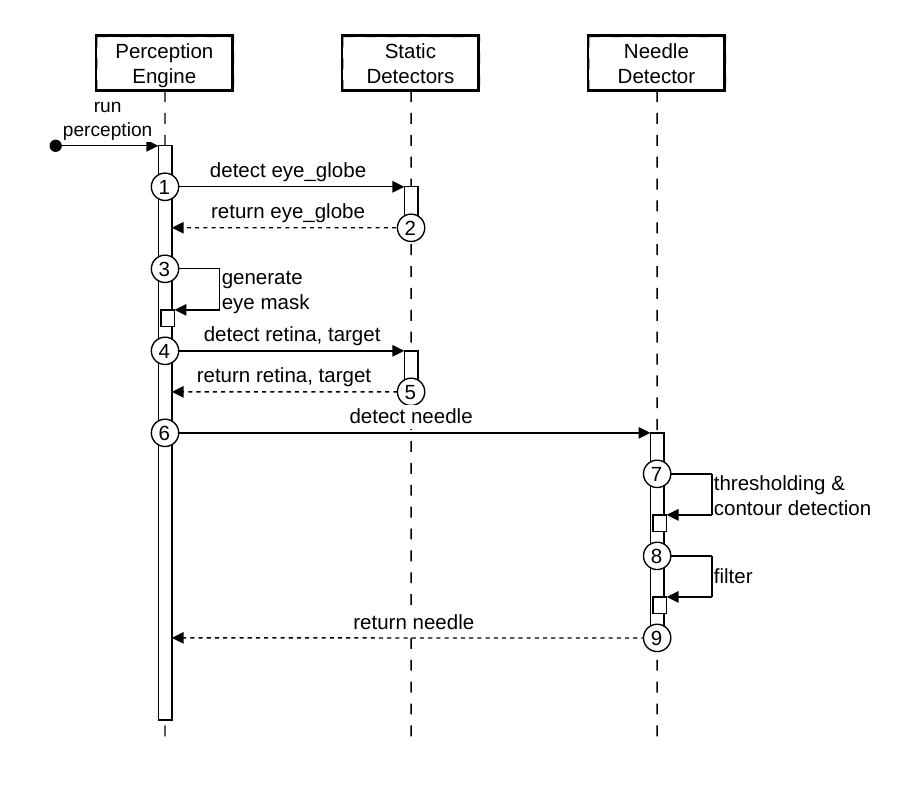}
    \caption{Perception Pipeline Sequence. The system captures the camera feed and sequentially triggers specific detectors for the eye globe (1), retina (4), and needle (6) to generate the geometric primitives required for Scene Graph construction.
    }
    \label{fig:pipeline}
\end{figure}
    \subsection{Perception Pipeline}
        The perception pipeline processes the video feed to track instruments relative to anatomical features inside the eye.

\begin{figure}[b]
    \centering
    \vspace{-10pt}
    \includegraphics[width=\linewidth]{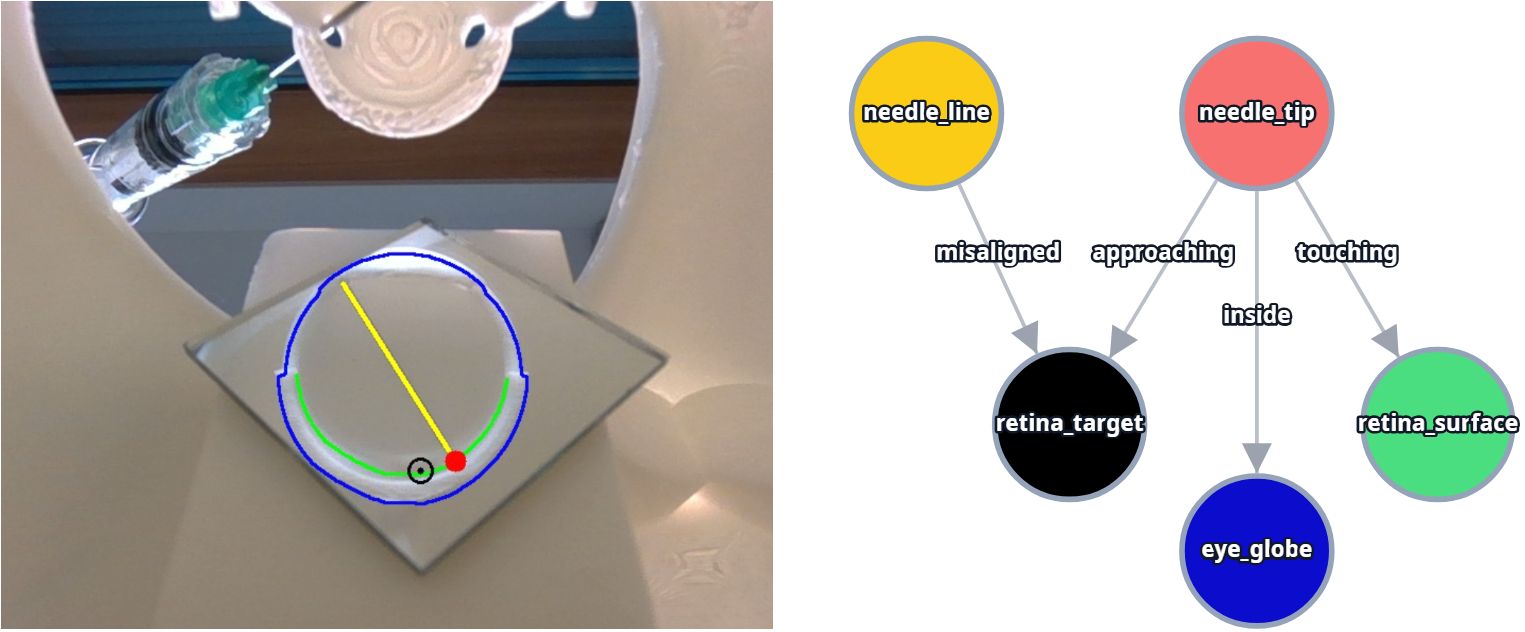}
    \caption{Surgical Scene Graph Generation. \textbf{Left:} The processed camera feed showing the surgical instrument interacting with the retina. \textbf{Right:} The corresponding Scene Graph representation, where nodes (circles) represent entities and edges (arrows) represent the active semantic relations used for haptic inference.
    % \textcolor{red}{HIER DAS BILD MIT DER MASKE vielleicht reinMACHEN!}
    }
    \label{fig:sg_vis}
\end{figure}

\begin{enumerate}
    \item \textbf{Anatomical Registration:} 
    The system is pre-calibrated using the static contours of immobile objects in the scene, such as the eye, the retina contours, and the target point (\Cref{fig:pipeline} (1-5)).
    % The system is pre-calibrated with the static contour of the eye globe. This is used to establish a region of interest, limiting future tracking to this area. The retinal surface curve, as well as the injection target point on it, is predefined. Since the eye model is completely static, pre-calibrated reference geometries are sufficient for our use case (see \cref{fig:pipeline} (1-5)).
    \item \textbf{Image Processing:} Rather than using a trained convolutional neural network to analyze the image, we simplified the process by applying adaptive Gaussian thresholding followed by contour detection (7) to accurately identify the needle shaft in the cross-sectional view. This design choice prioritizes \textit{safety}, \textit{explainability}, and \textit{low latency}. 
    % To increase the probability of robust detection, we first apply adaptive Gaussian thresholding, which computes local thresholds based on weighted neighborhood averages. 
    Block size and offset parameters were empirically tuned to counteract noise with instrument preservation in mind. 
    % Since the eye model is encapsulated inside the head mask, the system is also more resilient to varying lighting conditions across the OR.
    % making it resilient to varying lighting conditions across the surgical field. %The eye globe contour serves as a mask, filtering out irrelevant background regions before threshold computation.
    % \item \textbf{Instrument Tracking:} The visual tracking operates at the camera frame rate (30 Hz). After thresholding is applied to the image, we use a simple contour detection algorithm to identify the edge of an inserted surgical instrument. After that, the system fits a line through the surgical tool, extracting position and angle. 
    The analyzed result is shown on the left in \Cref{fig:sg_vis}. 
    %  counteract false positives, fixed instrument entry points are defined in the locations of the trocars. 
    % To mitigate other sensor noise and reflection artifacts, the needle movement is smoothened using an exponential moving average (8), achieving a desirable balance between stability and responsiveness.
    % $s_t = (1-\alpha)*s_{t-1} + \alpha * x_t$, where $s_t$ is the smoothed position and $\alpha \in (0,1]$ is a smoothing factor. 
    % A smaller $\alpha$  yields a smoother trajectory of the
    % needle line, but reduces responsiveness to sudden movements, while a larger $\alpha$ increases movement reactivity but also makes the estimate more sensitive to noise. 
    % Since the needle typically moves very slowly within the eye, a relatively smaller smoothing factor can achieve a desirable balance between stability and responsiveness.
    \item \textbf{Instrument Tracking:}     
    To mitigate sensor noise and artifacts, the needle movement is smoothed using an exponential moving average (8), achieving a desirable balance between stability and responsiveness.
\end{enumerate}

\begin{figure}
    \centering
    % \begin{subfigure}{.25\columnwidth}
    %     \centering
    %     \includegraphics[height=3cm]{pictures/vision/3.png}
    % \end{subfigure}\hfill
    % \begin{subfigure}{.25\columnwidth}
    %     \centering
    %     \includegraphics[height=3cm]{pictures/vision/2.png}
    % \end{subfigure}\hfill
    \begin{subfigure}{.5\columnwidth}
        \centering
        \includegraphics[height=4cm]{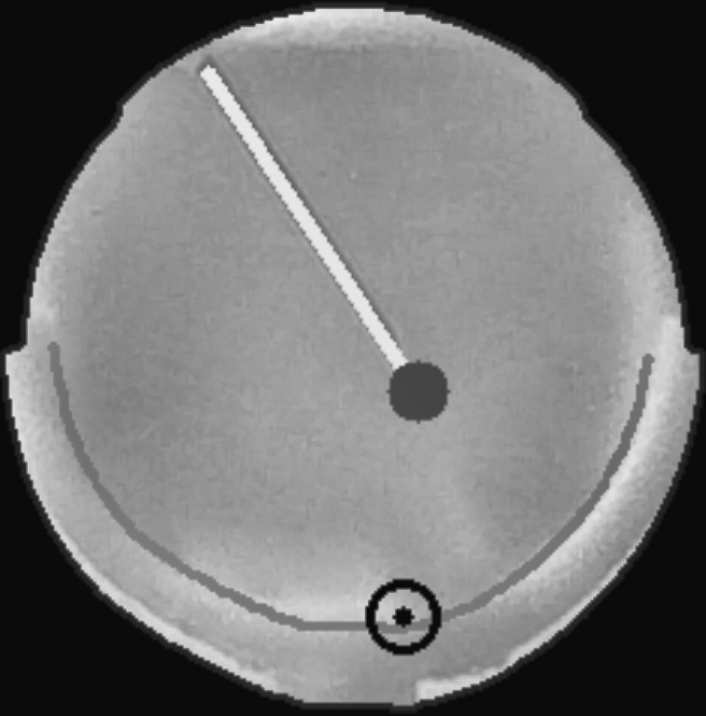}
    \end{subfigure}\hfill
    \begin{subfigure}{.5\columnwidth}
        \centering
        \includegraphics[height=4cm]{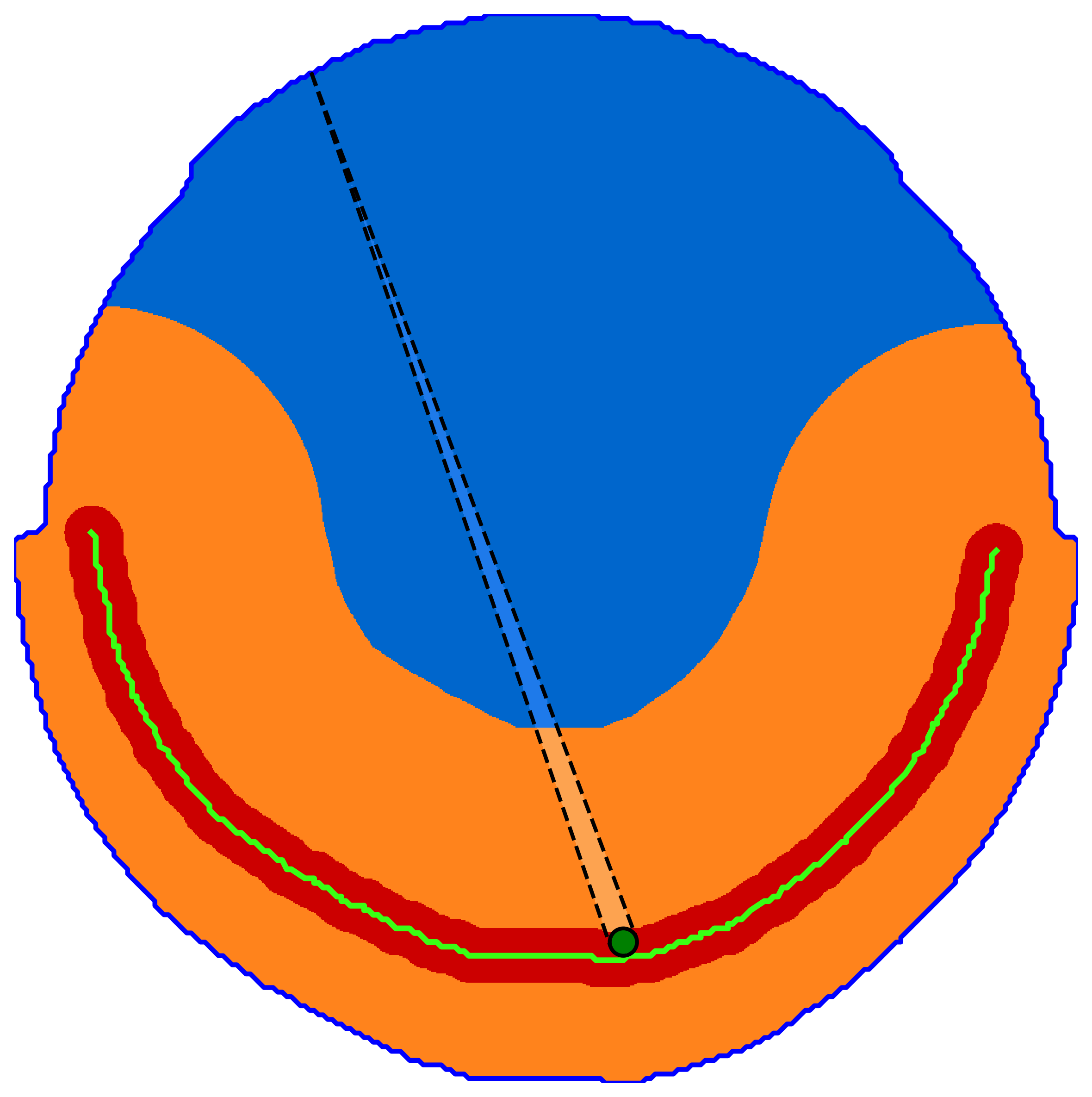}
    \end{subfigure}
    \caption{Needle detection and surgical state visualization.
    \textbf{Left:} Modified image with the detected needle contour overlaid after thresholding-based detection and ''iOCT-fication''. 
    \textbf{Right:} Visualization of Semantic Trigger Zones. 
    The system maps the needle's spatial position within these zones to discrete Scene Graph relations (e.g., approaching, aligned), which subsequently drive the inference engine logic. The cone represents optimal alignment to the target point.
    Colors match with \Cref{fig:state_transition_diagram} for surgical states that are indirectly assumed.}
    \vspace{-12pt}
    \label{fig:cross-sectional_view}
\end{figure}

To create similarity to OCT, the frame is slightly edited again after it has passed through the pipeline. Here, the eye phantom is cut out, zoomed in, and grayscaled as depicted on the left side of \Cref{fig:cross-sectional_view}.
This image is then displayed to the user as a substitute for a real iOCT feed. 

    \subsection{Surgical Scene Graph}
        To represent the surgical scene and bridge the gap between low-level camera data and the high-complexity surgical context, we utilized the concept of an SG as our semantic middleware for future decision-making. 
Formally, we define the SG at given time $t$ as the labeled, directed graph $G_t = (V,E,O,R,\phi)$, where $V$ is the set of nodes, $O$ is the set of entity categories, like retina or needle tip.
$R$ is the set of possible relations between nodes, like touching or inside.
$E \subseteq V \times R \times V$ is the set of directed and labeled edges, expressed by the triplet notation \textit{$\langle$subject, predicate, object$\rangle$} and the labeling function $\phi: V \rightarrow O$ assigning each node a category.

For every frame, the image is analyzed, and a corresponding SG is dynamically computed (\Cref{fig:sg_vis}, right).
The graph definition is shown in \Cref{tab:sg_model}.

To bridge the gap between continuous sensor data and discrete logic, we utilize the geometric primitives visualized in \Cref{fig:cross-sectional_view} (Right).
These spatial zones act as triggers for edge instantiation. 
For example, the pipeline detects if the needle tip falls within the defined Retinal Zone (distance $\leq 0.7mm$).
If true, a corresponding edge (\textit{$\langle$needle\_tip, touching, retina\_surface$\rangle$}) is added to the SG.
While the current SG generation primarily relies on geometric distances, the SG architecture serves as a critical abstraction layer. Unlike direct sensor-to-actuator mapping (e.g., hard-coded conditional loops), the SG decouples the perception pipeline from the feedback logic. This offers two key advantages: 
\begin{itemize}
    \item \textbf{Scalability:} New entities (e.g., \textit{hemorrhage}, \textit{surgical\_forceps}) or relations (e.g., \textit{tool-tool\_collision}) can be added to the graph schema without altering the downstream inference engine.
    \item \textbf{Context-Awareness:} The SG captures the procedure's global semantic state simply, enabling complex queries (e.g., ``Is the tool inside the eye AND touching the retina?'') that are robust to sensor noise compared to isolated geometric distance checks.
\end{itemize}

\begin{table}[h]
\centering
\caption{Definition of the Surgical Scene Graph with Entity Categories $O$, Semantic Relations $R$, and Geometric Trigger Thresholds. Additional nodes and relationships irrelevant to the feedback mechanism and study were omitted from the table. Thresholds were tuned to the phantom scale and expert feedback; clinical use requires dynamic adaptation to tissue deformation.}
\label{tab:sg_model}
\small
\setlength{\tabcolsep}{4pt}
% Adjusted column widths for IEEE Conference Column
\begin{tabularx}{\columnwidth}{
  p{1.8cm} % Group Column (Narrower)
  p{2.6cm} % Name Column
  >{\raggedright\arraybackslash}X % Description Column (Auto-fill)
}
\toprule
\textbf{Group} & \textbf{Category $O$} & \textbf{Description} \\
\midrule
\multirow{2}{=}{Instruments}
& \texttt{needle\_tip}  & 2D position of tip \\
& \texttt{needle\_line} & Line coords. of shaft \\
\midrule
\multirow{3}{=}{Anatomy}
& \texttt{retina\_surface} & Retina contour \\
& \texttt{retina\_target}  & Target point (2D) \\
& \texttt{eye\_globe}      & Eye globe contour \\
\midrule
\addlinespace % Adds a little gap between sections
\textbf{Pair} & \textbf{Relation $R$} & \textbf{Geometric Threshold} \\
\midrule
\multirow{2}{=}{Needle--Retina}
& \texttt{touching} & $d \leq 0.7$\,mm \\
& \texttt{approaching}     & $0.7 < d \leq 5.5$\,mm \\
\midrule
\multirow{4}{=}{Needle--Target}
& \texttt{touching} & Radial $d \leq 0.7$\,mm \\
& \texttt{approaching} & $0.7 < d \leq 5.5$\,mm \\
& \texttt{aligned}     & Angle $\theta \leq 1.7^\circ$ \\
& \texttt{misaligned}  & Angle $\theta > 1.7^\circ$ \\
\midrule
\multirow{2}{=}{Needle--Eye}
& \texttt{inside}  & Tip inside eye \\
& \texttt{outside} & Tip outside eye \\
\bottomrule
\end{tabularx}
\vspace{-12pt}
\end{table}
    \subsection{Rule-Based State Inference and Haptic Feedback}
        To keep the system predictable and explainable, we use a deterministic, rule-based inference engine rather than a probabilistic model.
The SG serves as the computational foundation of the engine: at each time step, the engine queries the SG topology $G_t$ to infer a discrete \textit{surgical state} $S_t$.
This state provides an explicit abstraction layer, making the resulting haptic command $H_t$ easier to interpret.

\begin{figure}[b!]
    \centering
    \vspace{-10pt}
    % \begin{subfigure}{.5\columnwidth}
    %     \centering
    %     \includegraphics[height=4cm]{pictures/states_mix.png}
    % \end{subfigure}\hfill
    % \begin{subfigure}{.5\columnwidth}
    %     \centering
    %     \includegraphics[height=4cm]{pictures/realStateSG.png}
    % \end{subfigure}
    \includegraphics[width=1.0\columnwidth]{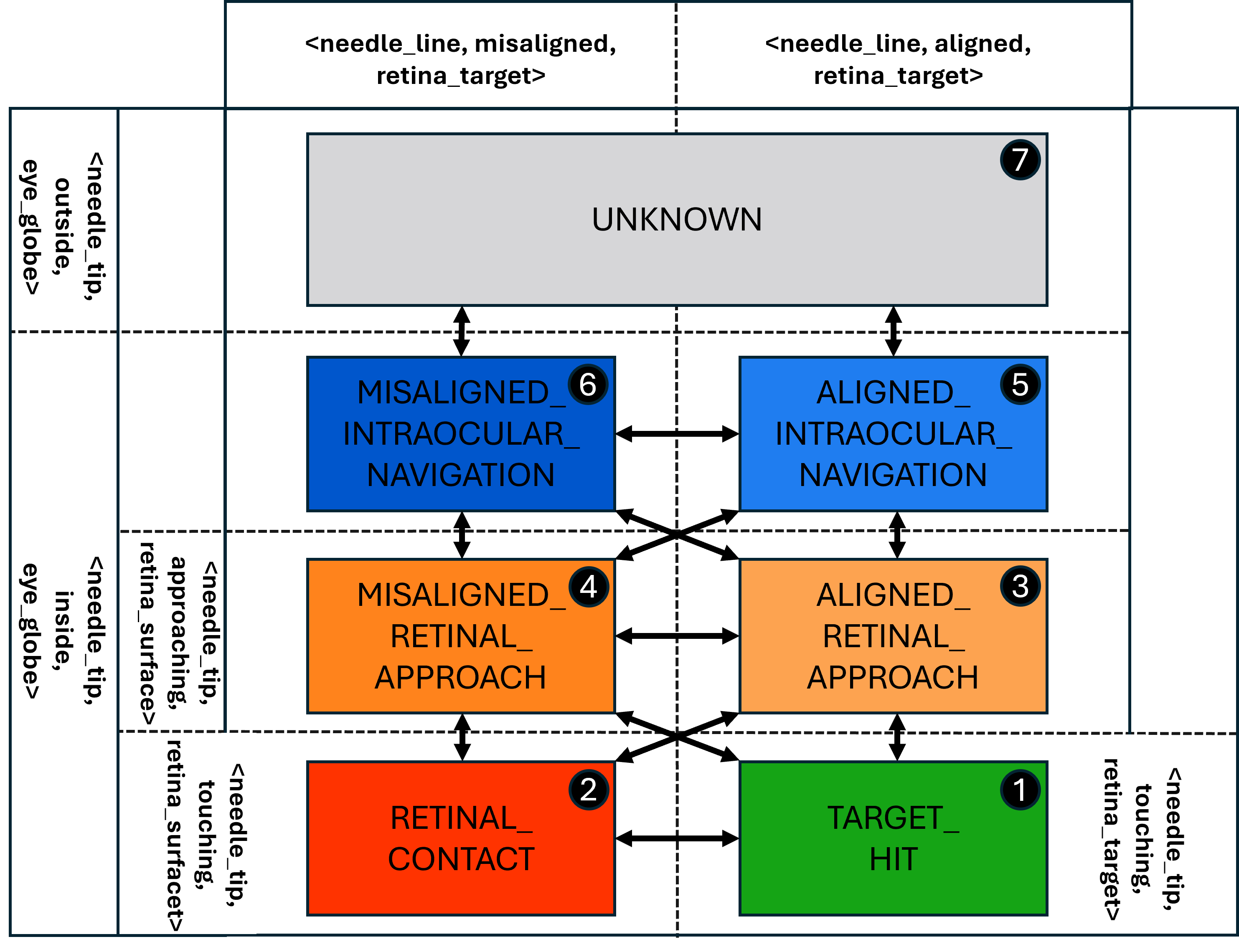}
    % \includesvg[width=0.80\columnwidth]{pictures/states.svg}
    \caption{
    State transition diagram.
    Misaligned states (left) trigger continuous vibration; aligned states (right) remain silent.
    All states are implicitly connected to \textit{Unknown}, representing failure of needle detection.
    Numbers denote rule priority (lower number = higher priority).
    }
    \label{fig:state_transition_diagram}
\end{figure}

\subsubsection{Surgical State Machine}
The inference engine operates as a finite-state machine with seven distinct surgical states, which are visualized in \Cref{fig:state_transition_diagram}.
The states are arranged into two parallel columns, divided by their semantic \textit{alignment} relation:
\begin{itemize}
    \item \textbf{Misaligned Column (Left):} This represents states where the surgical tools' trajectory deviates from the target point on the retina by an angle $\theta > 1.7^{\circ}$.
    \item \textbf{Aligned Column (Right):} This represents states where the tool is oriented toward the target point with $\theta \le 1.7^{\circ}$.
\end{itemize}
These seven surgical states are then mapped onto a specific haptic grammar.

\subsubsection{Haptic Grammar and Feedback Strategy}
We map these surgical states to haptic commands $H_t$ to implement an error-avoidance feedback strategy. The goal is to funnel the surgeon from the active feedback states into the silent states:

\begin{itemize}
    \item \textbf{Nominal Navigation (Silent):} When following the \textit{Aligned Path}, no haptic feedback is applied ($H_t = \textit{Off}$). This \textit{silence by default} minimizes sensory fatigue and overload, implicitly confirming that the current trajectory is safe to follow.
    
    \item \textbf{Active Guidance (Continuous):} States in the \textit{Misaligned Path} trigger a continuous low-frequency vibration ($H_t = \textit{Vib}_{low}$). This warns the surgeon that the current trajectory may result in retinal contact away from the target unless the instrument angle is corrected. The vibration persists until alignment towards the target point on the retina is corrected.
    
    \item \textbf{Critical Events (Red/Green):} High-priority events trigger distinct patterns. A collision with the retina on an unwanted position (\textit{Retinal Contact} state) triggers a rapid pulsing alarm ($H_t = \textit{Vib}_{pulse}$), while reaching the target (\textit{Target Hit} state) triggers a short double-burst success signal ($H_t = \textit{Vib}_{success}$).
\end{itemize}

% \begin{algorithm}[t]
% \caption{Haptic Feedback Inference Loop \textcolor{red}{Need changes to logic. Additionally, this has more or less the same content as Table 3.2 in your thesis, right? Why not use that?}}
% \label{alg:haptic_logic}
% \begin{algorithmic}[1]
% \Require Scene graph $G_t = (V, E)$
% \Ensure Haptic command $H_t \in \{\textit{Off}, \textit{Vib}_{low}, \textit{Vib}_{pulse}, \textit{Vib}_{success}\}$

% \State $H_t \gets \textit{Off}$ \Comment{Default state (Blue)}

% \If{$\langle needle, inside, eye \rangle \notin E$}
%     \State \Return $H_t$ \Comment{Outside eye}
% \EndIf

% \If{$\langle needle, tissue\_contact, retina \rangle \in E$}
%     \State $H_t \gets \textit{Vib}_{pulse}$
%     \Comment{\textbf{Red:} Collision alert}
%     \State \Return $H_t$
% \EndIf

% \If{$\langle needle, target\_hit, target \rangle \in E$}
%     \State $H_t \gets \textit{Vib}_{success}$
%     \Comment{\textbf{Green:} Target reached}
%     \State \Return $H_t$
% \EndIf

% \If{$\langle needle, approaching, target \rangle \in E$}
%     \If{$\langle needle, misaligned, target \rangle \in E$}
%         \State $H_t \gets \textit{Vib}_{low}$
%         \Comment{\textbf{Orange:} Negative Reinforcement (Misaligned)}
%     \Else
%         \State $H_t \gets \textit{Off}$
%         \Comment{\textbf{Blue:} Aligned approach}
%     \EndIf
% \Else
%     \State $H_t \gets \textit{Off}$
%     \Comment{Free navigation}
% \EndIf

% \State \Return $H_t$
% \end{algorithmic}
% \end{algorithm}

\subsubsection{Hierarchical Logic}
To ensure safety, transition logic inside the engine follows a strict priority hierarchy (\Cref{fig:state_transition_diagram}, numbers) implemented as a series of if-then-else rules.
Safety-critical conditions directly overrule task-related information.
For instance, the presence of the triplet $\langle\texttt{needle\_tip}, \texttt{touching}, \texttt{retina\_surface}\rangle$ in the SG immediately triggers the \textit{Retinal Contact} state, overriding any lower-priority states.
This hierarchical logic ensures that lower-priority guidance cues from the system never suppress more critical warnings.

\section{STUDY SETUP}
\begin{figure}[b]
    \centering
    \vspace{-7pt}
    \includegraphics[width=1.0\linewidth]{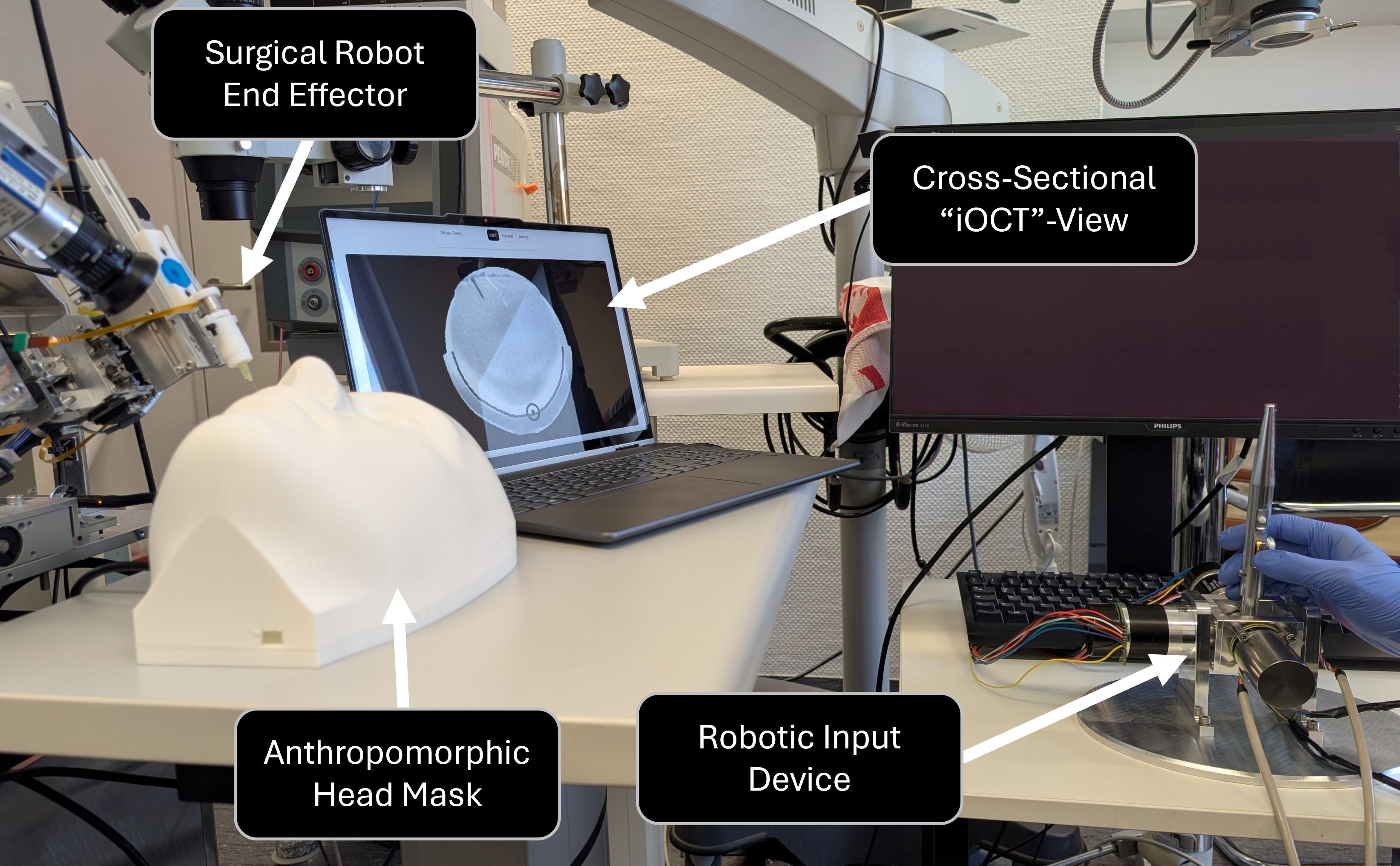}
    \caption{Study setup.}
    \label{fig:study_setup}
\end{figure}

To evaluate the proposed system, a user study was conducted with $N=16$ participants (university students and doctoral candidates, aged from 21 to 34).
The within-subjects study utilized a teleoperated robotic setup (\Cref{fig:study_setup}).
Participants controlled MAPS' surgical robot via a custom haptic input device.
The robot's end-effector was equipped with a surgical needle, translating movement of the input device into precise tool motion.
Feedback signals generated by the perception pipeline were sent in real-time to the input device, closing the teleoperation control loop.

Under a study protocol, participants performed needle insertion twice under different conditions: \textit{Visual Feedback Only} (Baseline) and \textit{Visual + Haptic Feedback} (Proposed). 
A run started in the \textit{Misaligned Intraocular Navigation} state and ended when the participant touched the target on the retina~(\textit{$\langle$needle\_tip, touching, retina\_target$\rangle$}).

The session began with a demographic survey and a standardized briefing on the subretinal injection procedure.
This ensured that all participants, regardless of their medical background, possessed a baseline understanding of the task and its constraints.
This was followed by a mandatory training session, during which participants familiarized themselves with the robotic controls and haptic cues while navigating unconstrainedly within the eye phantom. 
Thereafter, the users performed the tasks under both conditions, with the order randomized and the target point changing position on each trial to mitigate learning effects.
After completing each condition, participants filled out a System Usability Scale (SUS) questionnaire; the aggregated results are presented in \Cref{tab:results}.

To quantify the user's performance, different types of data were logged and evaluated.
We extracted the following quantitative metrics:

\begin{enumerate}
    \item \textbf{Task Completion Time:} The duration from needle detection inside the eye phantom (\textit{$\langle$needle\_tip, inside, eye\_globe$\rangle$}) to the moment it touches the target point successfully (\textit{$\langle$needle\_tip, touching, retina\_target$\rangle$})~(\Cref{tab:results}).
    \item \textbf{Alignment Angle:} The average angular deviation $\theta$ between needle shaft and trocar-target point line~(\Cref{tab:results}).
    \item \textbf{Surgical State Time Distribution:} 
    % To assess behavioral strategies for the two conditions, 
    Time spent in each surgical state~(\Cref{fig:condition_time_spent}).
    \item \textbf{Surgical State Order:} Order of traversed surgical states during the procedure (\Cref{fig:sequence_strategy}).
\end{enumerate}
Statistical significance was determined using a Wilcoxon signed-rank test with a significance level of $\alpha = 0.05$.
% The order of conditions was randomized to mitigate learning effects. 
% Before the participants began the conditions, a prior training phase was conducted, during which study participants were introduced to the robotic input device and had the opportunity to become accustomed to the controls.

\section{RESULTS}

\subsection{Task Efficiency}
Leveraging haptic feedback did not show any statistically significant change in task completion time~(\Cref{tab:results}).

\subsection{Safety and Error Reduction}
Safety was quantified by two primary metrics: the \textit{frequency} of unintended retinal contacts (touching retina without target contact) and the average \textit{Alignment Angle} $\theta$ maintained throughout the study.
% \textit{cumulative duration} of the time spent in misaligned surgical states close to the retina (\textit{Misaligned Retinal Approach}), and the alignment angle throughout the procedure.

Regarding critical safety incidents, all participants successfully evaded the negative \textit{Retinal Contact} state.
Nevertheless, an analysis of the needle trajectory shows a significant improvement in surgical precision.
As depicted in \Cref{tab:results}, the mean alignment error reduced from $7.9^\circ \pm 2.6^\circ$ to $6.8^\circ \pm 2.3^\circ$ ($p = 0.044$) in the feedback condition.
This indicates that haptic cues successfully nudged the user into a tighter safety corridor, minimizing angular deviation relative to the target point.
% Nevertheless, as detailed in \cref{fig:condition_time_spent}, when participants were conducting a run under vibrational feedback, they were quickly nudged into first aligning the needle and then approaching the retina.
% Without receiving feedback, the time spent in the \textit{Aligned Intraocular Navigation} state is nearly zero. 
% This indicates that nearly all participants in the baseline condition initially inserted the needle more deeply into the eye without prior alignment, approaching the retina with a potentially riskier trajectory. 
% In contrast, study runs with feedback often aligned themselves with the target and then inserted more deeply.

\begin{figure}
    \centering
    \vspace{5pt}
    \includegraphics[width=1.0\columnwidth]{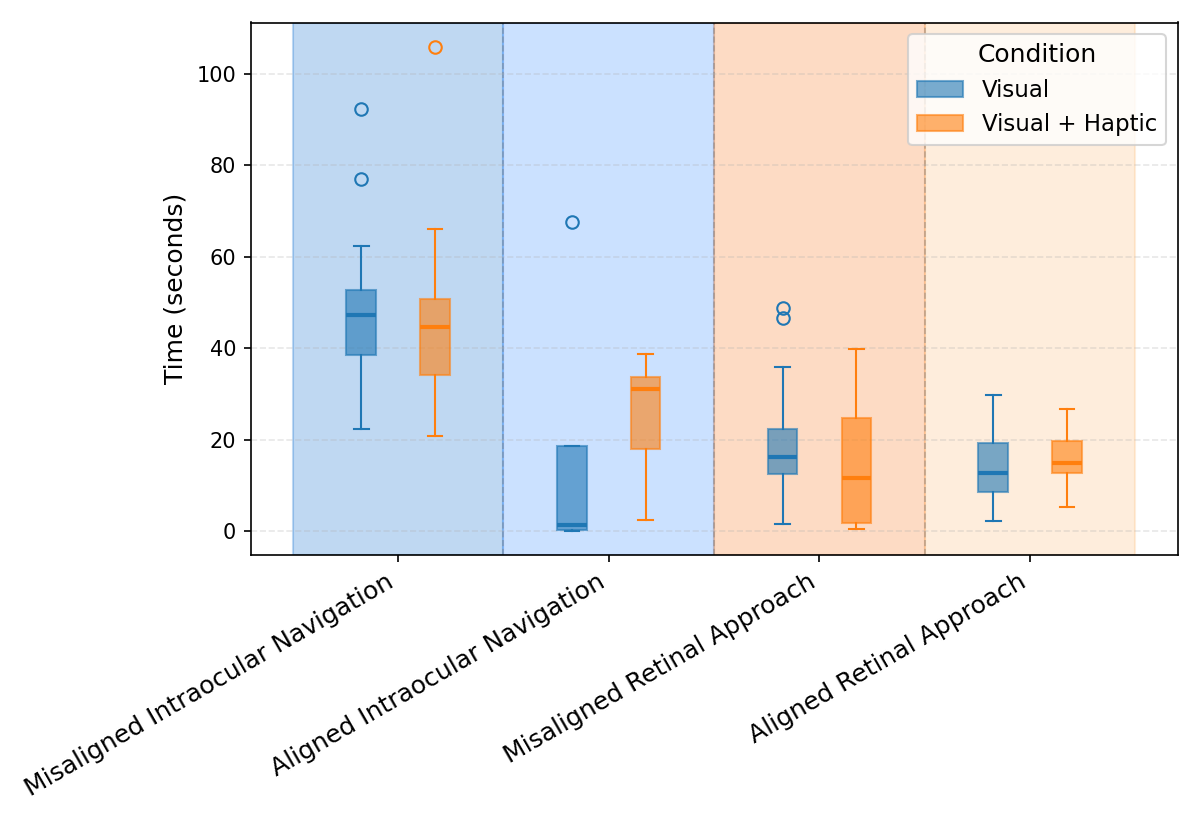}
    \caption{Surgical State Time Distribution. This shows a box plot comparing the time spent in each surgical state by participants, grouped by condition. Background colors match the \textit{Intraocular Navigation} and \textit{Retinal Approach} state colors from \Cref{fig:state_transition_diagram} and \Cref{fig:sequence_strategy}.
    }
    \vspace{-12pt}
    \label{fig:condition_time_spent}
\end{figure}

Interestingly, despite the improvements in the alignment angle, the duration of the \textit{Misaligned Retinal Approach} did not decrease significantly as hypothesized (\Cref{fig:condition_time_spent}).
This is likely caused by the geometric sensitivity of the perception pipeline: the angle estimation of the needle shaft to the target point is less robust at shallow insertion depths due to the shorter visible length.
Advancing the needle stabilizes and improves detection. 
This may cause slight deviations in the angle, prompting minor readjustments even after prior alignment, thereby accumulating time in the misaligned state even during a generally safer run.

% nearly \textcolor{red}{ $50\%$, dropping from $M_{V} = \text{INSERT\_COLLISIONS\_V}$ events to $M_{V+H} = \text{INSERT\_COLLISIONS\_VH}$ events ($p < 0.05$)}. Furthermore, the total time spent in dangerous contact with the retina was significantly reduced, indicating that when collisions did occur, participants reacted faster to the haptic cues to withdraw the instrument.

\begin{table}[b]
\centering
\vspace{-7pt}
\caption{Comparison of user performance metrics (Mean $\pm$ SD). * indicates statistical significance ($p < 0.05$).}
\label{tab:results}
\begin{tabular}{lccc}
\toprule
\textbf{Metric} & \textbf{Visual} & \textbf{Visual + Haptic} & \textbf{\textit{p}-value} \\
\midrule
Completion Time (s) 
& $87.3 \pm 20.4$ 
& $\mathbf{87.6 \pm 20.6}$ 
& $0.821$ \\

Alignment Angle ($^\circ$) 
& $7.9 \pm 2.6$ 
& $\mathbf{6.8 \pm 2.3}$ 
& $0.044^*$ \\

SUS Score (0--100) 
& $75.9 \pm 12.1$ 
& $\mathbf{81.9 \pm 14.3}$ 
& $0.015^*$ \\
\bottomrule
\end{tabular}
\end{table}

\subsection{Behavioral Strategy Analysis}
\begin{figure}[t]
    \centering
    \begin{subfigure}[t]{\columnwidth}
        \centering
        \includegraphics[width=\linewidth]{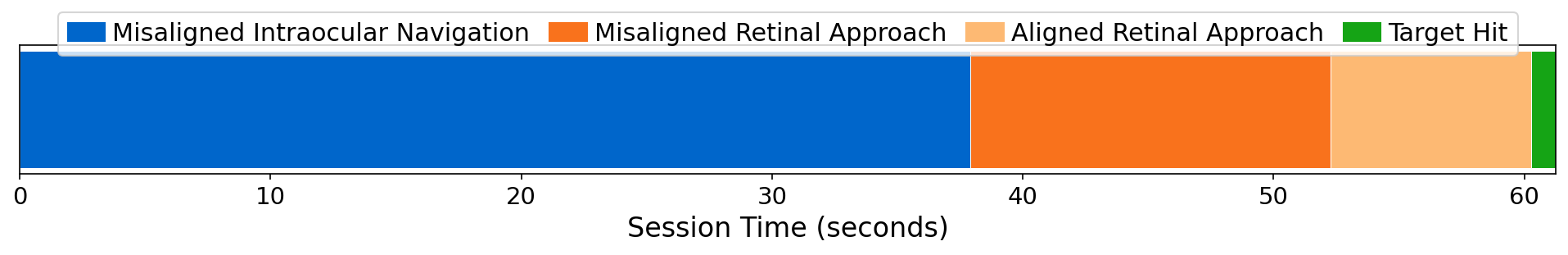}
        \caption{Visual only}
        \label{fig:sequence_strategy_a}
    \end{subfigure}

    \vspace{0.2em}

    \begin{subfigure}[t]{\columnwidth}
        \centering
        \includegraphics[width=\linewidth]{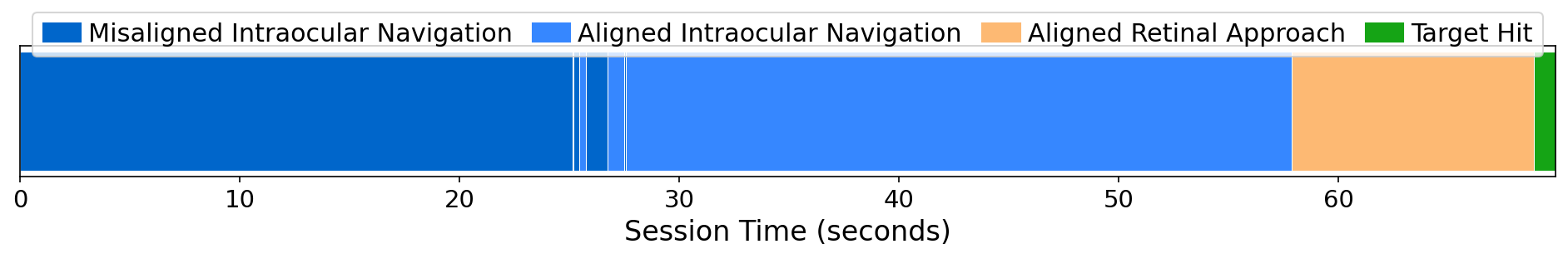}
        \caption{Haptic feedback}
        \label{fig:sequence_strategy_b}
    \end{subfigure}
    \caption{Sequence strategy diagrams illustrating the surgical states traversed during the procedure.  \textbf{Top (Visual Only):} Participants often approach the retina misaligned, leading to later angle corrections.  \textbf{Bottom (Haptic Feedback):} Participants align the tool early (light blue state) before approaching the retina, demonstrating an "Align-then-Approach" strategy.
    }
    \vspace{-12pt}
    \label{fig:sequence_strategy}
\end{figure}

To further investigate \textit{how} the improved alignment was achieved, the time distribution of surgical states (\Cref{fig:condition_time_spent}) and the dominant state transition sequence for both conditions (\Cref{fig:sequence_strategy}) were analyzed.
The data shows a distinct difference in movement strategy when individuals are under the influence of haptic feedback (\Cref{fig:sequence_strategy}).
In the \textit{Visual Only} condition, the median time spent in \textit{Aligned Intraocular Navigation} was nearly zero.
This indicates that the needle was first inserted more deeply into the eye before its angle was adjusted (\Cref{fig:sequence_strategy_a}).
By skipping initial alignment, the retina was approached with a riskier trajectory, necessitating corrective movements in close proximity to the sensitive tissue and potentially increasing the risk of iatrogenic injury. 
% To further investigate how sequence of surgical states was analyzed to understand how haptic feedback influenced the participants' control strategies. \Cref{fig:sequence_strategy} visualizes the dominant state transitions for both conditions.

% In the \textit{Visual Only} condition (\cref{fig:sequence_strategy_a}), participants frequently entered the \textit{Misaligned Retinal Approach} state, necessitating corrective movements in close proximity to the sensitive retinal tissue. 
% Although \textit{Retinal Contacts} were successfully avoided in the study, this behavior indicates a riskier trajectory where critical angulation adjustments are performed after the surgical tool has already been inserted close to the retina, possibly increasing the risk for iatrogenic injury.
% This could result in oscillatory behavior between \textit{Approach} and \textit{Retinal Contact}.

In contrast, the \textit{Visual + Haptic} condition  (\Cref{fig:sequence_strategy_b}) successfully promoted an ``Align-then-Approach'' strategy. 
The vibration cues for misalignment ($H_t = \text{Vib}_{low}$) encouraged participants to correct their trajectory early in the \textit{Misaligned Intraocular Navigation} phase. 
Consequently, they approached the retina primarily through the \textit{Aligned Intraocular Navigation} state, minimizing the risk of accidental contact during the final approach. 
This shift in behavioral strategy validates the context-aware nature of the SG-driven feedback.

\subsection{Usability and Cognitive Load}
To ensure that the added haptic modality did not introduce excessive distraction, we administered a System Usability Scale (SUS) questionnaire. As shown in \Cref{tab:results}, the system received a mean SUS score of $81.9$, which corresponds to a rating of ``Good''~\cite{bangor_determining_2009}. 
A significant increase of nearly $8\%$ in the score counters the concern that additional haptic feedback might overwhelm a user.
Instead, the results demonstrate that the guidance improved usability without introducing additional complexity.
% , effectively lowering the mental workload required to navigate safely.
% The difference between conditions was found to be statistically significant, indicating that the haptic feedback provided effective guidance with a positive impact on the user's perception of the system's usability and hints towards less perceived workload. 

% \begin{figure}[h]
%     \centering
%     \includegraphics[width=0.8\columnwidth]{pictures/sus_scores.png}
%     \caption{Distribution of System Usability Scale (SUS) scores. High scores in the Haptic condition indicate that the feedback was intuitive and did not distract the user.}
%     \label{fig:sus_scores}
% \end{figure}

\section{DISCUSSION AND CONCLUSION}
Standard feedback systems often rely on rigid, hard-coded thresholds. By contrast, our SG-driven approach provides a modular framework. Although the current experiment focuses on a specific injection task, the underlying graph structure enables the seamless integration of additional surgical contexts, such as anatomical constraints or workflow phases, without requiring the re-engineering of the control architecture.
Consequently, we propose that the community should explore SGs not merely for analysis, but also as a semantic middleware for active robotic control.
A significant observation was the user response to the 'Misaligned' vibration. Rather than ignoring the cue, participants actively altered their trajectory to silence the haptic signal. This suggests that the system successfully employed negative reinforcement cues to promote safer surgical behavior, effectively creating a 'soft' virtual fixture without active motor restriction. 
Currently, alignment is only binary. In future work, we will explore continuous guidance and address study limitations, specifically the sample size ($N=16$), absence of tissue deformation, and imaging noise, by validating on more realistic models and clinical OCT data.
% Ultimately, this shows promise as an intraoperative safety net and surgical training tool.
%Currently, the alignment is only binary; either it is aligned or not. In future work, we intend to use continuous guidance, e.g., vibration intensity increasing with the amount of misalignment. 
Ultimately, this ability to implicitly guide movement highlights the system's potential not only as an intraoperative safety net but also as a tool for skill acquisition in surgical training.
% \textcolor{red}{The main reviewer criticism we might receive is that a camera is not iOCT and thus our vision pipeline will not work with it. While true, this is the exact reason we are using the scene graph - to abstract away the sensor modality. However, we should emphasize it, probably in this section. Write something like "While our perception pipeline utilizes standard computer vision on a camera feed, the Scene Graph acts as a modality-agnostic interface. Whether the nodes (e.g., 'Retina', 'Tool') are detected via intensity thresholding on a camera or U-Net segmentation on real iOCT, the downstream haptic inference engine remains unchanged."}.

% \section{ACKNOWLEDGEMENT}
% The authors acknowledge the use of Google's Gemini model to assist in structuring and drafting the manuscript.
% All generated content was reviewed, edited, and finalized by the authors.

%%%%%%%%%%%%%%%%%%%%%%%%%%%%%%%%%%%%%%%%%%%%%%%%%%%%%%%%%%%%%%%%%%%%%%%%%%%%%%%%

% This tells LaTeX to use the IEEEtran.bst file you uploaded
\bibliographystyle{IEEEtran} 

% This loads the standard abbreviations (IEEEabrv.bib) AND your own references (references.bib)
\bibliography{IEEEabrv,references}

\end{document}